\documentclass{bmvc2k}
\usepackage{booktabs}
\usepackage{hyperref}
\title{Multimodal Hate Detection Using Dual-Stream Graph Neural Networks}

\addauthor{Jiangbei Yue}{j.yue@exeter.ac.uk}{1}
\addauthor{Shuonan Yang}{sy446@exeter.ac.uk}{1}
\addauthor{Tailin Chen}{tailin1009@gmail.com}{1}
\addauthor{Jianbo Jiao}{j.jiao@bham.ac.uk}{2}
\addauthor{Zeyu Fu}{Z.Fu@exeter.ac.uk}{1}

\addinstitution{Multimodal Intelligence Lab, Department of Computer Science\\
University of Exeter\\
Exeter, UK}
\addinstitution{School of Computer Science\\
University of Birmingham\\
Birmingham, UK}
\runninghead{Jiangbei Yue et al.}{MultiHateGNN for Hateful Video Classification}

\begin{document}

\maketitle

\begin{abstract}
Hateful videos present serious risks to online safety and real-world well-being, necessitating effective detection methods. Although multimodal classification approaches integrating information from several modalities outperform unimodal ones, they typically neglect that even minimal hateful content defines a video's category. Specifically, they generally treat all content uniformly, instead of emphasizing the hateful components. Additionally, existing multimodal methods cannot systematically capture structured information in videos, limiting the effectiveness of multimodal fusion. To address these limitations, we propose a novel multimodal dual-stream graph neural network model. It constructs an instance graph by separating the given video into several instances to extract instance-level features. Then, a complementary weight graph assigns importance weights to these features, highlighting hateful instances. Importance weights and instance features are combined to generate video labels. Our model employs a graph-based framework to systematically model structured relationships within and across modalities. Extensive experiments on public datasets show that our model is state-of-the-art in hateful video classification and has strong explainability. Code is available: \url{https://github.com/Multimodal-Intelligence-Lab-MIL/MultiHateGNN}.

{\color{red}Disclaimer: This paper contains sensitive content that may be disturbing to some readers.}
\end{abstract}

%-------------------------------------------------------------------------
\section{Introduction}
\label{sec:intro}
With the rapid development of various video platforms, the volume of video content has increased dramatically, necessitating automated systems to effectively detect and address hateful content \cite{das2023hatemm,wang2024multihateclip}. As a result, hateful video classification has attracted considerable attention. Existing classification approaches generally fall into two categories: unimodal and multimodal. Unimodal models \cite{das2023hatemm,wang2024multihateclip} focus on a single modality, such as visual frames or text, and often overlook hateful cues present in other modalities, leading to limited accuracy. To overcome this, multimodal models \cite{das2023hatemm,wang2024multihateclip,zhang2024enhanced} have been proposed, which integrate information across different modalities and significantly outperform unimodal approaches.

However, existing multimodal methods often fail to fully consider the asymmetric nature of hateful videos, namely, that the presence of any amount of hateful content is sufficient for the video to be classified as hateful. Ideally, classification models should highlight the hateful parts and diminish the influence of non-hateful ones in hateful videos. Yet, current methods typically integrate all components equally, resulting in poor performance. Specifically, this simple equal treatment typically has difficulty in classifying correctly videos with sparse hateful content because of the interference of a large volume of non-hateful content. Moreover, while multimodal fusion has shown promise, the exploration of how best to fuse information remains limited. Existing methods generally ignore the systematic modelling of crucial structured information within videos, such as intra-modal temporal structure and inter-modal relational structure.

To address these two key challenges, we propose MultiHateGNN, a novel multimodal dual-stream graph neural network. This graph-based framework systematically models both intra-modal and inter-modal structured information, enabling the extraction of representative multimodal features. Specifically, we divide each input video into $N$ segments and use pre-trained models to extract features from multiple modalities, visual frames, audio, and text, for each segment. We then construct two types of graphs using the extracted features: an instance graph and a weight graph. To reduce interference from non-hateful content, the instance graph partitions the video into $K$ instances, each comprising $N/K$ segments, and builds a subgraph for each instance. These subgraphs are processed by a shared graph neural network (GNN) \cite{velivckovic2018graph,wu2020comprehensive} to extract multimodal features. The weight graph is fed into another GNN to estimate the importance weights of the subgraphs. Our model emphasizes hateful instances and suppresses non-hateful instances through importance weights to achieve robust classification performance. The features from the subgraphs are aggregated, weighted according to their importance before being passed to a classifier for final label prediction.

Our MultiHateGNN model leverages structured information in videos through graph neural networks. Additionally, our method employs dual-stream graphs to mitigate the interference of non-hateful content in hateful videos. We demonstrate that our model achieves the state-of-the-art performance on public datasets. Formally, our contributions are as follows:
\vspace{-1mm}

\begin{itemize}
    \item We propose a novel multimodal dual-stream graph neural network, MultiHateGNN, to classify hateful videos effectively. To our best knowledge, this is the first work to utilize graph-based modelling to explore hateful video classification.
    \item We propose a new dual-stream graph architecture to emphasize crucial content for video classification.
    \item We demonstrate that our model achieves the state-of-the-art performance on the widely used public dataset HateMM \cite{das2023hatemm} and MultiHateClip \cite{wang2024multihateclip}. Our model can also provide the explanation for predictions through estimated instance importance weights.  
\end{itemize}

\section{Related Work}
The dissemination of hateful content threatens the harmonious digital space. There are a large number of studies on the detection of hateful content \cite{caselli2020hatebert,koutlis2023memefier,zhang2024enhanced}. Researchers first focused on hateful speech classification, which relies on a single textual modality. Existing methods for hateful speech classification are generally divided into traditional and deep learning methods. Traditional approaches \cite{chen2012detecting,warner2012detecting,waseem2016hateful,davidson2017automated} tend to rely on hand-crafted features. Deep learning methods \cite{gamback2017using,founta2019unified,caselli2020hatebert} automatically learn meaningful features from data using neural networks, significantly reducing the need for manual feature engineering.

As the Internet continues to evolve, hateful content is no longer limited to textual forms. Hateful memes, comprising an image and embedded text, have become increasingly prevalent across digital platforms \cite{lu2024towards}. Consequently, the detection of hateful memes has garnered significant research attention \cite{kiela2020hateful,pramanick2021momenta,lu2024towards}. For example, MemeFier \cite{koutlis2023memefier} is a model based on a dual-stage modality fusion framework for hateful meme classification. The model encodes images and text through pre-trained models, respectively. These encoded features are aligned in the first stage and then fed into a transformer \cite{vaswani2017attention} to capture inter-modal relationships and output effective representations in the second stage. Furthermore, MemeFier incorporates external knowledge and background image caption supervision to enhance its classification performance. Ji \textit{et al.} \cite{ji2023identifying} introduced a new prompt-based method to detect hateful memes. The method generates captions and attributes from the visual content of the memes. The generated textual descriptions are appended to the text embedded in the memes to obtain the pure text input, which is fed into a pre-trained language model to identify hateful memes. 

Recently, research on multimodal hateful content detection has been extended to hateful videos involving the visual, audio, and text modality. The complexity of video data poses challenges in data collection and classification. Das \textit{et al.} \cite{das2023hatemm} constructed a valuable multimodal video dataset, comprising 1083 videos and annotated them as hate or non-hate. Additionally, they presented a multimodal classification model, which employs pre-trained models to extract visual, audio, and text features. After projecting these features into embeddings with equal dimensionality through corresponding neural networks, the model concatenates the embeddings for the classification. Later, Zhang \textit{et al.} \cite{zhang2024enhanced} proposed CMFusion, a powerful multimodal classification model. CMFusion also first extracts visual, audio, and text features via pre-trained models. Subsequently, CMFusion introduces a temporal cross-attention module to align visual and audio features. Eventually, the new channel-wise and modality-wise modules are designed to fuse features across three modalities and produce informative representations for the classification. To address problem of data scarcity in hateful video classification, Wang \textit{et al.} \cite{wang2025cross} presented a novel method utilizing meme data to improve classification performance. Specifically, they first re-annotated the meme data to ensure the label consistency between meme and video data. Then, they fine-tuned a vision-language model, such as LLaMA-3.2-11B \cite{grattafiori2024llama}  or LLaVA-NeXT-Video-7B \cite{zhang2024llava}, by using the video dataset and re-annotated meme datasets to detect hateful videos. However, these multimodal models fail to highlight crucial hateful content and capture structured information in videos, which significantly limits their classification performance. Our MultiHateGNN introduces dual-stream graphs to emphasize hateful instances and model intra- and inter-modal structured correlations.

\section{Method}
\begin{figure*}
    \centering
    \includegraphics[width=0.9\linewidth]{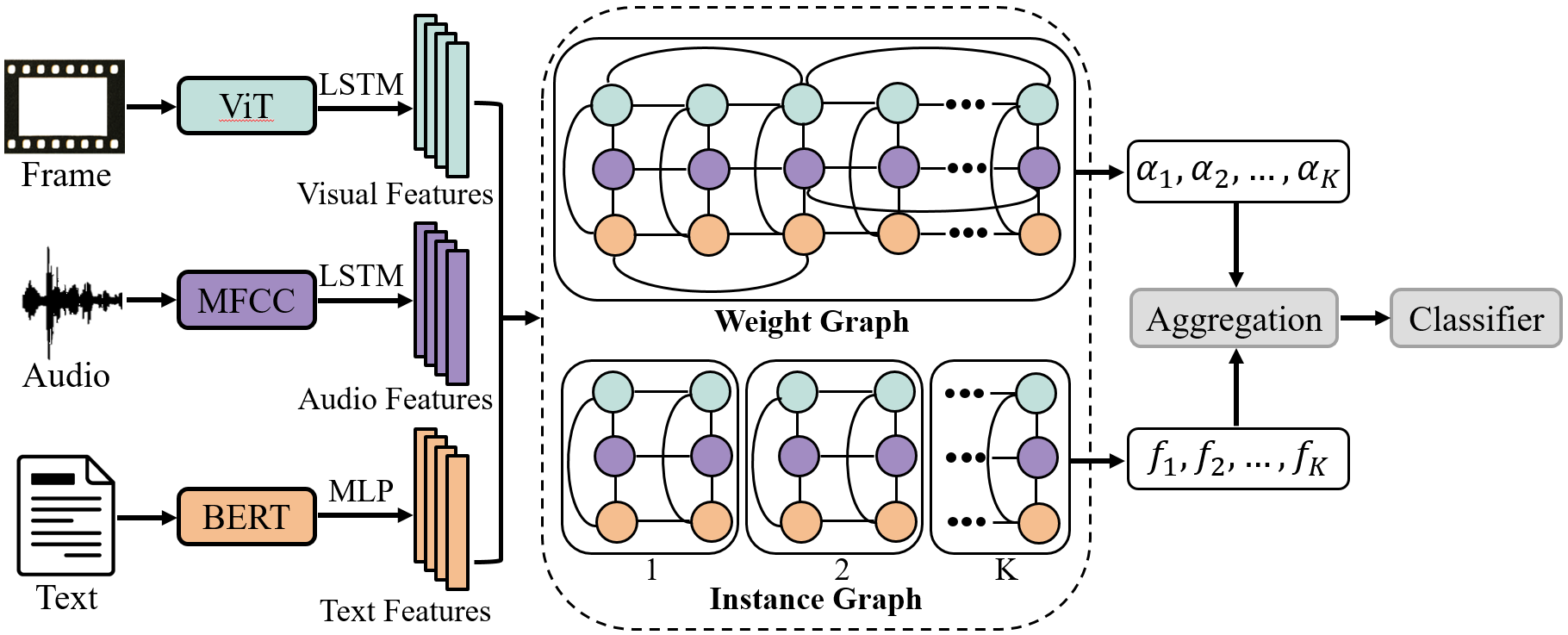}
    \caption{Overview of Our Model. We divide an input video into $N$ segments and extract three types of segment-level features, \textit{i.e.} visual, audio and text features. The weight graph and instance graph are built based on segment-level features. These two graphs are fed into corresponding GNNs to yield instance features $\{ f_i \}_{i=1}^{K}$ and instance importance weights $\{ \alpha_i \}_{i=1}^{K}$, respectively. Finally, after aggregating $\{ f_i \}_{i=1}^{K}$ and $\{ \alpha_i \}_{i=1}^{K}$, a classifier predicts video labels.  
     }
    \label{fig:overview}
\end{figure*}

\subsection{Overall Model}
Formally, the hateful video classification task is to classify a video as hate (y = 1) or non-hate (y = 0), aligning with previous work. We present an overview of our proposed model in Fig. \ref{fig:overview}. Inspired by previous methods, our model adopts the way to cut the given video into $N$ segments to capture video content efficiently. We extract a key frame, audio, and text from each segment to obtain raw multimodal information, where text is the transcript from audio. We then employ pre-trained models to extract modality-specific representations, which are passed through corresponding projection layers to produce segment-level features of uniform dimensionality. This process yields three categories of segment-level features: visual, audio, and text features.  We construct the instance graph and the weight graph based on the segment-level features. The instance graph consists of $K$ sub-graphs corresponding to different instances and is fed into a GNN to produce instance features $\{ f_i \}_{i=1}^{K}$. The weight graph is passed through another GNN to obtain importance weights $\{ \alpha_i \}_{i=1}^{K}$ for instances. The aggregation layer integrates instance features $\{ f_i \}_{i=1}^{K}$ and importance weights $\{ \alpha_i \}_{i=1}^{K}$ to output aggregated features. Finally, a classifier based on neural networks takes the aggregated features as input and predicts video labels.

\subsection{Segment-Level Feature Extraction}
We extract the first frame of each segment and sequentially feed them into the pre-trained Vision Transformer \cite{dosovitskiy2020image} (ViT) to obtain $N$ 768-dimensional features. These features are fed into a Long Short Term Memory (LSTM) as a project layer to yield $N$ $D$-dimensional visual features, where $D$ denotes the dimension of node representations in the following graphs. The audio in each segment is also extracted. We employ Mel Frequency Cepstral Coefficients \cite{muda2010voice} (MFCC) to extract the 40-dimensional feature for each segment-level audio. These features are also fed into an LSTM to obtain $N$ $D$-dimensional audio features. Then, we generate a paragraph of text by collecting all text associated with the segment for each segment to retain the semantic integrity. To be specific, we utilise Whisper \cite{radford2023robust} to transcribe the entire audio in a video into text, sentence by sentence. A sentence is incorporated into the paragraph for a segment if the time spans of the sentence and the segment overlap. We feed all paragraphs sequentially into BERT \cite{devlin2019bert} to extract $N$ 768-dimensional features. These features are fed into a multi-layer perceptron (MLP) as a project layer to output $N$ $D$-dimensional text features. We finally have three segment-level features with the same shape $[N,D]$.

\subsection{Weight Graph}
We first establish the weight graph based on extracted segment-level features. The weight graph $\mathcal{G}_W=\{\mathcal{V}_W, \mathcal{E}_W\}$ consists of a node set $\mathcal{V}_W$ and an edge set $\mathcal{E}_W$. We construct a node for each modality in every segment. This results in 3$N$ nodes, including $N$ visual nodes, $N$ audio nodes and $N$ text nodes. Visual/Audio/Text nodes have corresponding segment-level visual/audio/text features as their node representations. Subsequently, we construct undirected edges to model intra- and inter-modal correlations. We first connect temporally adjacent nodes within the same modality. Additionally, we adopt the $\epsilon-$graph principle \cite{tenenbaum2000global}, which facilitates the discovery of structured information. Specifically, two nodes in the same modality are connected if the distance between their representations is smaller than the given threshold $\epsilon$. Here, we use the cosine distance:
\begin{equation}
    dist(v_i, v_j) = 1 - \frac{v_i \cdot v_j}{\| v_i \|_2 \| v_j \|_2},
\end{equation}
where $v_i$ and $v_j$ are representations for nodes i and j, respectively. Eventually, nodes at the same timestamp across the three modalities are connected in a pairwise manner to model and capture inter-modal relationships.

The weight graph is fed into a GNN to learn intra- and inter-modal relationships, and the GNN outputs new node representations. We employ an MLP to take each new node representation as input and yield node importance scores $\{ \alpha_j^v \}_{j=1}^{N}$, $\{ \alpha_j^a \}_{j=1}^{N}$, and $\{ \alpha_j^t \}_{j=1}^{N}$. Then, the node importance scores within the same modality are fed into a softmax layer to obtain node importance weights $\{ \hat{\alpha}_j^v \}_{j=1}^{N}$, $\{ \hat{\alpha}_j^a \}_{j=1}^{N}$, and $\{ \hat{\alpha}_j^t \}_{j=1}^{N}$. Finally, we can calculate the instance importance weight by:
\begin{equation}
    \alpha_i = \frac{1}{3}\sum_{l \in \Omega_i}  \hat{\alpha}_l^v +  \hat{\alpha}_l^a +  \hat{\alpha}_l^t,
\end{equation}
where $\Omega_i$ denotes the set of all segments covered by the ith instance. 

\subsection{Instance Graph}
To establish the instance graph, we divide temporally the input video into $K$ non-overlapping instances, where each instance contains the same number of segments. We construct sub-graphs for each instance in the same way used to build the weight graph. Subsequently, a shared GNN is used to process all sub-graphs one by one and outputs the new learned node representations. We aggregate the learned node representations to obtain the instance features in every subgraph. Specifically, we first average the learned node representation within the same modality:
\begin{equation}
     f_i^{*} = \frac{1}{M} \sum_{p=1}^{M_i} f_i^{*,p}, \,\, *=v, a, \text{or} \, t,    
\end{equation}
where $M_i = N/K$ denotes the number of nodes in the ith sub-graph and $*$ represents the modality. We then concatenate $f_i^{v}, f_i^{a}, f_i^{t}$ to produce the instance feature from the ith sub-graph as: 
\begin{align}
    f_i = [f_i^v \, \| \, f_i^a \, \| \, f_i^t].
\end{align}
Following a similar way, instance features $\{ f_i \}_{i=1}^{K}$ are calculated. All subgraphs form the complete instance graph. We opt for a simple arithmetic average to compute $f_i^*$, rather than a weighted average based on the node importance weights. This is because we assume that the contributions of learned node representations within a subgraph are relatively uniform, and we expect the weight graph to emphasize differences at the instance level. Instead of using methods such as averaging for multimodal fusion, we concatenate $f_i^{v}$, $f_i^{a}$, and $ f_i^{t}$ to attain instance features, as concatenation preserves the complete multimodal information without losing modality-specific features.

We note that the introduction of the instance graph can significantly reduce the interference of non-hateful content on classifying hateful videos. Specifically, we usually have several non-hateful instances and multiple hateful instances given a video. A substantial volume of non-hateful content is placed in non-hateful instances. Hateful instances tend to have higher ratios of hateful content than the original video. Therefore, we reduce the interference of non-hateful content in hateful videos when extracting instance features $\{ f_i \}_{i=1}^{K}$. Furthermore, we can highlight important instance features by aggregating importance weights $\{ \alpha_i \}_{i=1}^{K}$ and instance features $\{ f_i \}_{i=1}^{K}$.

\subsection{Prediction and Loss Function}
The aggregation layer calculates the classification feature $f$ as:
\begin{equation}
    f = \sum_{i=1}^K \alpha_i f_i,
\end{equation}
where $\alpha_i$ denotes the importance weight and $f_i$ is the instance feature. Subsequently, the classifier consisting of an MLP and a softmax layer takes the classification feature $f$ as input to predict the video one-hot label:
\begin{equation}
    \hat{h} = \phi(MLP(f)),
\end{equation}
where $\hat{h} = [\hat{h}_0, \hat{h}_1]$ is a 2-dimensional vector, $\hat{h}_0$ and $\hat{h}_1$ represent the probabilities of non-hate and hate labels, respectively, and $\phi$ denotes the softmax layer. 

To train our model, we use one-hot encoding to denote video labels, \textit{i.e.} $[1,0]$ and $[0,1]$ represent non-hate (y=0) and hate (y=1), respectively. Then, we employ the cross-entropy loss function \cite{zhang2018generalized}:
\begin{equation}
    \mathcal{L}_{CE} = -\sum_{c=1}^2 h_c \log(\hat{h}_c),
\end{equation}
where $h=[h_0, h_1]$ is the ground-truth one-hot label. 

We introduce implementation details in the supplementary material (SM), including the determination of the number of segments $N$ and the number of instances $K$, the architetures of the neural networks used in our model, \textit{etc.}. 

\section{Experiments}
\subsection{Dataset and Metrics}
We use the most popular public datasets in hateful video detection, HateMM \cite{das2023hatemm} and MultiHateClip (MHC) \cite{wang2024multihateclip}, to evaluate the classification performance of our model. \textbf{HateMM} is a multimodal dataset widely used for hateful video detection, integrating visual, audio, and textual modalities. It contains annotated video clips that span various forms of hate speech, enabling the study of complex cross-modal cues and context. HateMM consists of 652 non-hateful videos and 431 hateful videos. We follow the 5-fold stratified cross-validation protocol to evaluate our model and other baselines, aligning with previous research. The standard dataset splitting in \cite{das2023hatemm} is employed, where we have the 70\%/10\%/20\% split for training, validation, and testing, respectively.

\textbf{MHC} is the newest multimodal dataset for hateful video detection, comprising 1,000 English-language videos. These videos are classified into three categories: non-hateful (662), offensive (256), and hateful (82). MHC also combines the offensive and hateful categories for the binary classification, resulting in 662 non-hateful videos and 338 hateful videos. Following the experimental setup in MHC, we run each model five times and report the averaged results, where we also have the 70\%/10\%/20\% split for training, validation, and testing, respectively.

Following existing studies \cite{das2023hatemm,zhang2024enhanced,liang2024fusion}, we use metrics of Accuracy, F1-score, Precision, and Recall to evaluate our model and all baseline methods. These metrics jointly provide a comprehensive assessment of the models' classification performance. We use the hate label (y=1) as the positive label to calculate F1-score, Precision, and Recall. 

\subsection{Quantitative Results}
\begin{table}[tb]
    \centering
    \begin{tabular}{c|c|c|c|c}
    \toprule
        Model & Accuracy & F1-score & Precision & Recall \\
        \midrule
        ViT & 0.693 & 0.601 & 0.623 & 0.589 \\
        MFCC & 0.682  & 0.622 & 0.602 & 0.651 \\
        BERT & 0.706 & 0.630 &  0.646 & 0.622 \\
        GPT-4o & 0.777 & 0.755 & 0.671 & \textbf{0.863} \\
        HateMM & 0.805 & 0.753 & 0.765 & 0.751 \\
        CMFusion & 0.799 & 0.739 & 0.763 & 0.719 \\
        Ours & \textbf{0.821} & \textbf{0.771} & \textbf{0.798} & 0.754 \\ 
    \bottomrule
    \end{tabular}
    \vspace{1mm}
    \caption{The Classification Performance on the HateMM dataset. The best results are shown in bold.}
    \label{tab:cls_hateMM}
\end{table}

To demonstrate the effectiveness of our model, we compare it against unimodal models and the state-of-the-art multimodal approaches. The experimental results on the HateMM dataset are shown in Table \ref{tab:cls_hateMM}.ViT, MFCC, and BERT serve as unimodal baselines from \cite{das2023hatemm} for the visual, audio, and textual modality, respectively. HateMM and CMFusion \cite{zhang2024enhanced} are the state-of-the-art multimodal models, where HateMM is the benchmark multimodal method used in the HateMM dataset \cite{das2023hatemm}. For fair comparison, our model and multimodal baselines adopt the same unimodal features for fusion, \textit{e.g.} ViT, MFCC and BERT on the HateMM dataset. Additionally, we introduce a large language model baseline, GPT-4o \cite{hurst2024gpt}, which determines the video label based on its transcripts. More details about the GPT-4o baseline can be found in the SM. We note that multimodal models significantly outperform unimodal models, demonstrating the importance of incorporating multimodal information. Moreover, our model achieves state-of-the-art performance across all metrics except for Recall, particularly improving the accuracy from 80.5\% to 82.1\%. We note that only the GPT-4o outperforms our model in Recall. However, it has a significantly low precision, demonstrating that it classifies many or even most samples as positive, regardless of whether they truly are. In summary, our model achieves the state-of-the-art classification performance on the HateMM dataset.   

\begin{table}[tb]
    \centering
    \begin{tabular}{c|c|c|c|c}
    \toprule
        Model & Accuracy & F1-score & Precision & Recall \\
        \midrule
        ViViT & 0.73 & 0.68 & \textbf{0.86} & 0.57 \\
        MFCC & 0.54  & 0.36 & 0.40 & 0.33 \\
        mBERT & 0.57 & 0.52 &  0.68 & 0.42 \\
        GPT-4o & 0.68 & 0.29 & 0.57 & 0.19 \\
        MHC & 0.75 & 0.67 & 0.77 & 0.61 \\
        CMFusion & 0.73 & 0.72 & 0.72 & 0.73 \\
        Ours & \textbf{0.78} & \textbf{0.77} & 0.80 & \textbf{0.77} \\ 
    \bottomrule
    \end{tabular}
    \vspace{1mm}
     \caption{The Classification Performance on the MHC dataset. The best results are shown in bold.}
    \label{tab:cls_mhc}
\end{table}

Additionally, we present the experimental results on the MHC dataset in Table \ref{tab:cls_mhc}. ViViT (Video Vision Transformer) \cite{arnab2021vivit}, MFCC, mBERT \cite{mozafari2019bert} are three unimodal baselines from \cite{wang2024multihateclip} for the visual, audio, and textual modality, respectively. MHC is the benchmark multimodal method used in the MHC dataset \cite{wang2024multihateclip}. Similar to the HateMM dataset, multimodal baselines and our model utilize the same unimodal features, including ViViT, MFCC and mBERT. Our model outperforms other baselines on all metrics except for Precision. Only one baseline, ViViT, achieves a better precision than our model. However, this baseline has poor performance in Recall, showing that it is not sensitive enough to detect all positive samples. We note that our model significantly improves the previous best accuracy and F1-score by 3\% and 5\% on the MHC dataset. Therefore, our model also is state-of-the-art in classification performance on the MHC dataset. Moreover, the outstanding performance in diverse settings further demonstrates that our model is robust. We owe the excellent classification performance of our model to the construction of the two-stream graph, which highlights crucial instances and captures vital structured information.  

\subsection{Qualitative Results}
\begin{figure}
    \centering
    \includegraphics[width=\linewidth]{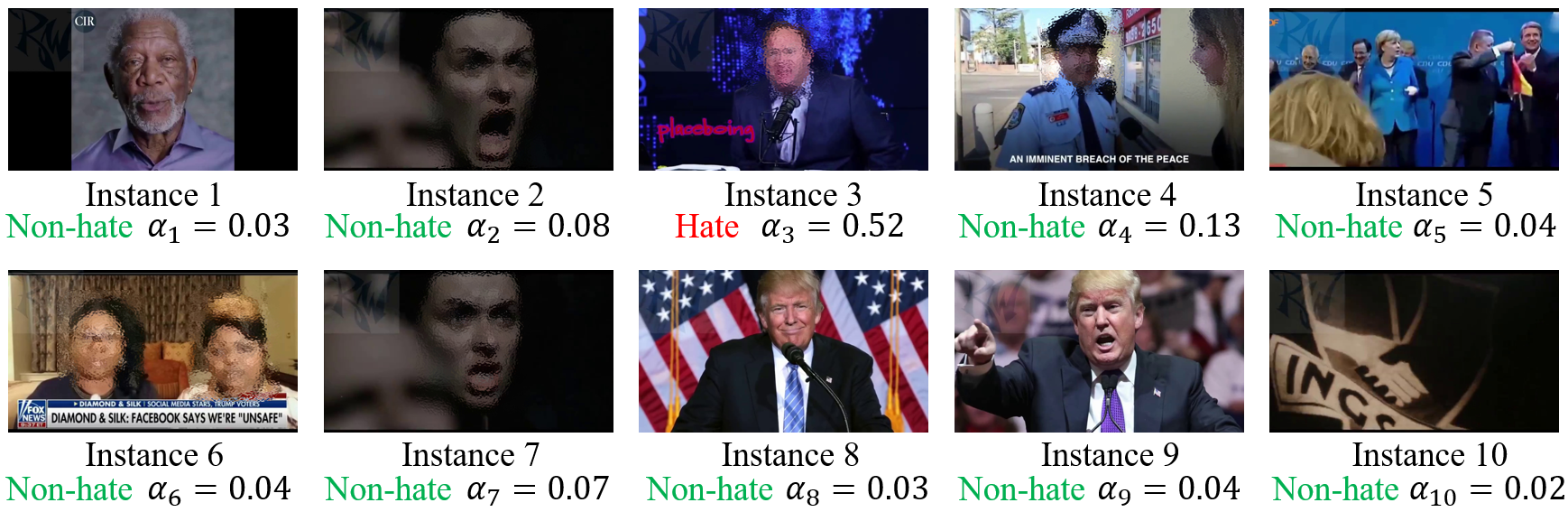}
    \caption{The Visualization of Model Prediction. Every instance is annotated as hate or non-hate. $\{ \alpha_j \}_{j=1}^{10}$ denote the instance weights.}
    \label{fig:qualitative_2}
\end{figure}

We show the visualization of our model prediction in Fig. \ref{fig:qualitative_2}, where a video from the HateMM dataset is divided into 10 instances and each instance is classified as hate if it involves hateful content. Classifying this video correctly is challenging because only one instance contains hateful content, while the remaining nine are non-hateful. Existing methods often struggle with this scenario due to the overwhelming presence of non-hateful content, which interferes with accurate classification. However, our model can accurately highlight the crucial hate instance 3 by the associated weight $\alpha_3 = 0.52$ to predict the correct label.

\textbf{Explainability}. Different from pure black-box existing methods, our model is capable of providing plausible explanations for the model predictions. For example, our model mainly considers the hateful instance features of the instance 3 to produce the hate label for the video in Fig. \ref{fig:qualitative_2}. We can see that these instance importance weights are also beneficial for the localization of hateful content.

\subsection{Ablation Study}
\begin{table}[tb]
    \centering
    \begin{tabular}{c|c|c|c|c}
    \toprule
        Model & Accuracy & F1-score & Precision & Recall \\
        \midrule
        No Graph & 0.792 & 0.742  & 0.759  & 0.727  \\
        Only Instance Graph & 0.798 & 0.746 & 0.767 & 0.726\\
        Only Wight Graph & 0.813 & 0.756  & 0.789 & 0.738 \\
        Full Model & \textbf{0.821} & \textbf{0.771} & \textbf{0.798} & \textbf{0.754} \\ 
    \bottomrule
    \end{tabular}
    \vspace{1mm}
    \caption{Ablation Study on the HateMM dataset. The best results are shown in bold.}
    \vspace{-1mm}
    \label{tab:ablation}
\end{table}

The dual-stream graph plays a crucial role in our model. To evaluate its effectiveness, we introduce three baselines and conduct an ablation study. The first one, No Graph, maintains the extraction of visual/audio/text features but removes the dual-stream graph. No Graph then averages the features within the same modality. Finally, the averaged features across three modalities are concatenated and fed into the MLP classifier. The second baseline only retains the instance graph. The instance features are aggregated without any weight based on the arithmetic average, which is passed through the MLP classifier for classification. The last baseline, only weight graph. After estimating node importance weights, we calculate the weighted average within the same modality. We also concatenate three averaged features and feed it into the MLP classifier.

We demonstrate the results of the ablation study in Table \ref{tab:ablation}. The worst performance of No Graph proves the necessity of introducing graphs. Only the Weight Graph is the best baseline, which shows that highlighting crucial content is effective. Eventually, our full model achieves the state-of-the-art classification performance by combining the instance graph with the weight graph.   

\section{Conclusion}
In this paper, we propose MultiHateGNN, a new multimodal two-stream graph neural network for hateful video classification. We employ the weight graph and the instance graph to highlight crucial content and model structured information in videos. Our proposed model achieves state-of-the-art classification performance on the public dataset across diverse settings. In the future, we would like to extend our model through multi-scale analysis to localize hateful content, which could improve the efficiency of current content moderation. We will also explore the combination of a large language model and graph neural networks in hateful video detection.       

\section*{Acknowledgements}
The research work was supported by the Alan Turing Institute and DSO National Laboratories Framework Grant Funding.

\bibliography{egbib}
\end{document}